\documentclass[conference]{IEEEtran}
\IEEEoverridecommandlockouts
\usepackage{amsmath,amsfonts}
\usepackage{algorithmic}
\usepackage{algorithm}
\usepackage{array}
\usepackage[caption=false,font=normalsize,labelfont=sf,textfont=sf]{subfig}
\usepackage{textcomp}
\usepackage{stfloats}
\usepackage{url}
\usepackage{verbatim}
\usepackage{graphicx}
\usepackage{cite}
\usepackage{url}

\def\BibTeX{{\rm B\kern-.05em{\sc i\kern-.025em b}\kern-.08em
    T\kern-.1667em\lower.7ex\hbox{E}\kern-.125emX}}

\begin{document}

\title{Ship in Sight: Diffusion Models for Ship-Image Super Resolution
\thanks{This work was partly supported by ``Ricerca e innovazione nel Lazio - incentivi per i dottorati di innovazione per le imprese e per la PA - L.R. 13/2008" of Regione Lazio, Project ``Deep Learning Generativo nel Dominio Ipercomplesso per Applicazioni di Intelligenza Artificiale ad Alta Efficienza Energetica", under grant number 21027NP000000136, and by the European Union under the Italian National Recovery and Resilience Plan (NRRP) of NextGenerationEU, ``Rome Technopole" (CUP B83C22002820006)—Flagship Project 5: ``Digital Transition through AESA radar technology, quantum cryptography and quantum communications". }
}

\author{\IEEEauthorblockN{Luigi Sigillo$^{\ast, \dagger}$, Riccardo Fosco Gramaccioni$^{\ast, \dagger}$ , Alessandro Nicolosi$^{\dagger}$ and Danilo Comminiello$^{\ast}$}
        \IEEEauthorblockN{\textsuperscript{*}\textit{Dept. Information Engineering, Electronics and Telecommunications (DIET), Sapienza University of Rome, Italy}}
        \IEEEauthorblockN{$^\dagger$\textit{Leonardo Labs, Via Tiburtina Km. 12.400, Rome 00156, Italy}\\
Email: luigi.sigillo@uniroma1.it.}
}




\maketitle

\begin{abstract}
In recent years, remarkable advancements have been achieved in the field of image generation, primarily driven by the escalating demand for high-quality outcomes across various image generation subtasks, such as inpainting, denoising, and super resolution. A major effort is devoted to exploring the application of super-resolution techniques to enhance the quality of low-resolution images. In this context, our method explores in depth the problem of ship image super resolution, which is crucial for coastal and port surveillance. We investigate the opportunity given by the growing interest in text-to-image diffusion models, taking advantage of the prior knowledge that such foundation models have already learned. In particular, we present a diffusion-model-based architecture that leverages text conditioning during training while being class-aware, to best preserve the crucial details of the ships during the generation of the super-resoluted image.
Since the specificity of this task and the scarcity availability of off-the-shelf data, we also introduce a large labeled ship dataset scraped from online ship images, mostly from  ShipSpotting\footnote{\url{www.shipspotting.com}} website. 
Our method achieves more robust results than other deep learning models previously employed for super resolution, as proven by the multiple experiments performed.
Moreover, we investigate how this model can benefit downstream tasks, such as classification and object detection, thus emphasizing practical implementation in a real-world scenario. Experimental results show flexibility, reliability, and impressive performance of the proposed framework over state-of-the-art methods for different tasks. The code is available at: \url{https://github.com/LuigiSigillo/ShipinSight}
\end{abstract}


\section{Introduction}
\begin{figure}[!t]
\centering
\includegraphics[width=\linewidth]{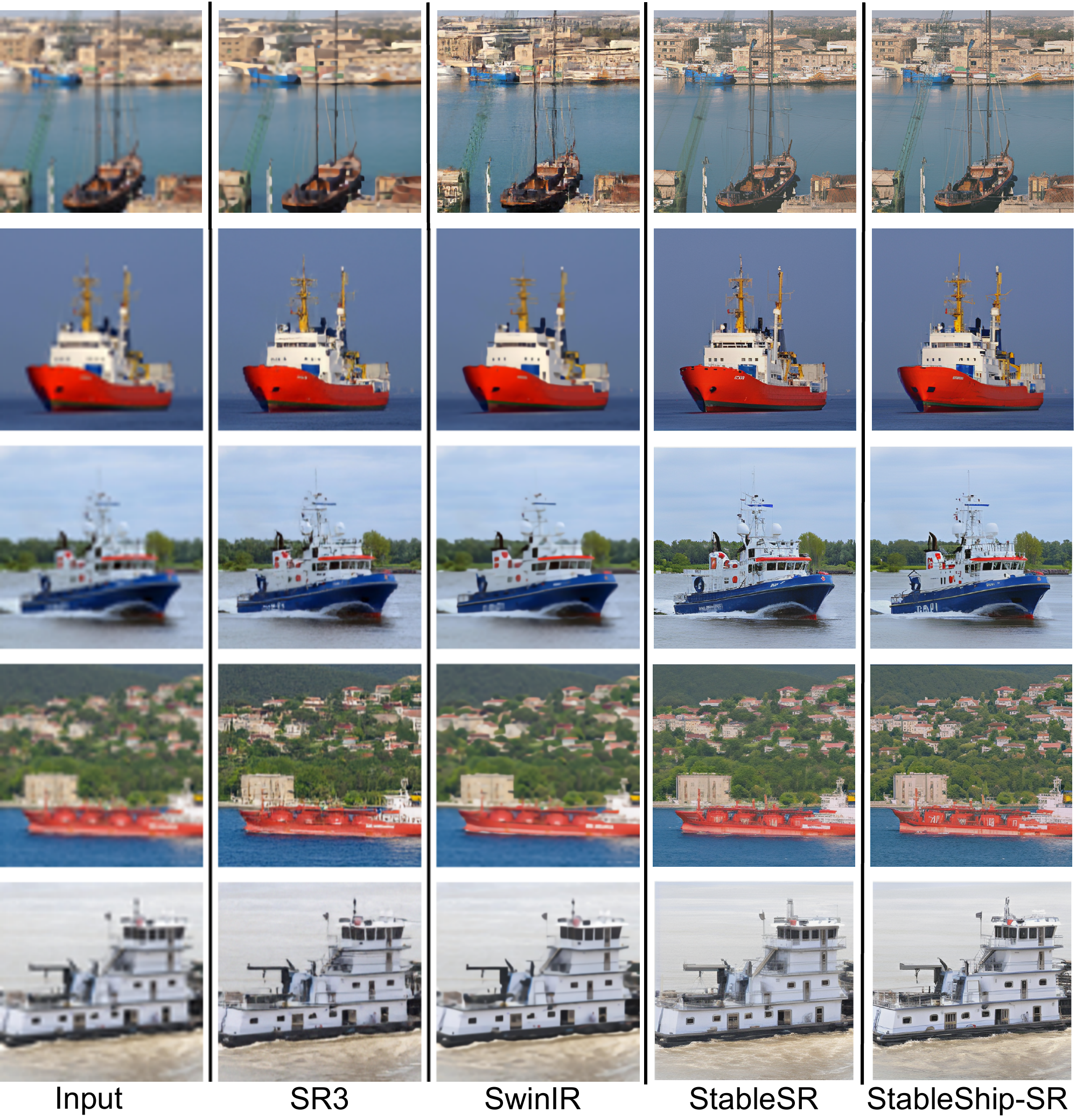}
\caption{Comparative sample images showcasing the super-resolution capabilities of different models. The visual assessments highlight variations in image quality, clarity, and detail enhancement achieved by each respective model in the upscale process. Ours is in the last column.}
\label{fig:sample_all}
\end{figure}

Our application domain deviates slightly from the conventional ones where super-resolution models are typically evaluated. Traditionally, super resolution primarily focuses on natural or face images \cite{ledig2017photorealistic, Menon2020PULSESP}, which present intriguing challenges due to the brain familiarity with these subjects. 
However, super resolution is a task of pivotal importance in various other domains, including maritime surveillance. Indeed, this domain presents several challenges \cite{Karakus2019ShipWD, Yang2018PositionDA, Maritime_Vessel_Classification, 10285968}. In this domain, considerable factors, such as long distances, limited sensor capabilities, adverse weather conditions, and the inherent mobility of ships prevent the acquisition of high-quality images.
With the enhancement of low-resolution ship images, potential defects, anomalies, or critical details that might have been missed in the original pictures can be revealed. This capability enables the identification and addressing of issues more effectively, enhancing quality control measures, and overall operational efficiency.
Those aspects are critically important for several applications, including ship detection \cite{seaships}, classification \cite{8455679}, and tracking \cite{9024119}. One of the reasons why those images are important for those downstream tasks is that there exists a wide variety of ship categories, each possessing distinct characteristics. For example, consider a cargo ship capable of transporting different types of containers or being empty. Furthermore, ships may exhibit diverse small and highly detailed elements that require enhancement, including portholes, railings, crests, text, and even armaments. 
The super-resolution process must yield high-quality results to mitigate the risks of mistaking for instance a container for a cannon, or vice versa. This process is not as easy as with natural and face images, where smooth edges are expected, indeed ships comprise steel or other components with sharp and well-defined features. Preserving these characteristics in the enhanced ship images is essential for minimizing the artifacts generated in the super-resolution process as much as possible.
Taken together, all these factors underscore the considerable challenges inherent in this domain, showing that ship-image super resolution holds significant relevance for industrial application problems. 


Indeed generative models in recent years, especially diffusion models \cite{Song2020DenoisingDI}, have emerged as promising solutions to address those problems by effectively enhancing low-resolution images while preserving their visual fidelity \cite{Wang2019DeepLF}. 
The use of generative models, which can hallucinate missing details from a low-quality source image, has proven to greatly benefit image-to-image translation problems \cite{Wan2020OldPR, Yeh2016SemanticII, Ulyanov2017DeepIP, Zamir2021RestormerET, stawgan}. 
The unsupervised nature of those kinds of models allows for the enhancement of low-resolution images without relying on large annotated datasets \cite{ledig2017photorealistic, Wang2018ESRGANES, Menon2020PULSESP}. In literature, different papers showed how we can take advantage of the diffusion process to perform single-image super resolution \cite{LI202247, Gao_2023_CVPR} even with different kinds of inputs such as hyperspectral images \cite{10353979} showing a broad interest in different fields of application.
Moreover, it is possible to guide the generation process using different kinds of conditioning including text, semantic maps, or like in our case a low-resolution version of an image. In literature, there are different approaches to condition diffusion models. One of the most popular is the classifier-free guidance \cite{pmlr-v162-nichol22a} where the conditioning element is removed from the objective function with some probability only for certain iterations of the training. The counterpart of this method is the classifier guidance approach \cite{Dhariwal2021DiffusionMB} where instead they use the gradients of an external classifier to condition the learning of the denoising process.

In this paper, we present our model StableShip-SR which uses as well a pre-trained classifier, but instead of relying on a fixed class prediction, it exploits this information as an embedding in a latent space together with the embeddings of the low-resolution image. The conditioning of the generation in this way is performed at different scales during the denoising process. Indeed the architecture for the conditioning is based on a class- and time- aware encoder which improves the performances of the pre-trained foundation models Stable Diffusion \cite{Rombach2021HighResolutionIS}. We exploit the prior knowledge present in the state-of-the-art text-to-image generative models, to improve and propose a novel ship-image super-resolution architecture. 
Our method shows generalizability across different datasets. For instance the results of a zero-shot super resolution on Seaships \cite{seaships} show remarkable improvements in the downstream task of detection and classification using pre-trained state-of-the-art models \cite{Wang_2023_CVPR}.  
StableShip-SR can benefit different areas of interest such as maritime navigation, defense, and environmental monitoring, facilitating advancements in ship-based systems and enabling improved decision-making processes. Given the difficulty in accessing accurate datasets for solving this specific problem, we have subsequently developed a dedicated dataset for training of our model.


Accordingly, our main contributions are as follows:
\begin{enumerate}
    \item We introduce a large labeled dataset of ship images comprehending more than 20 classes.
    \item We improve the state-of-the-art ship image super resolution by exploiting a latent diffusion model and introducing a novel class- and time- aware encoder.
    \item We conducted exhaustive experiments on the dataset we introduced comparing our model with some of the main deep learning models for image super resolution.
    \item We conducted ablation studies on different downstream tasks on diverse datasets comparing the improvement in performances using our StableShip-SR and other models for image super resolution.

\end{enumerate}
The paper is structured as below: in Section~\ref{sec:background} we introduce the theoretical elements upon which this work is based, the task we want to solve, and its application domain, discussing its uses and potential issues. In Section~\ref{sec:method} we introduce the proposed method StableShip-SR, while in Section~\ref{sec:dataset} we address the problem of the dataset. We present the obtained results in Section~\ref{sec:experiment} and finally in Section~\ref{sec:conclusion} we expose the results and draw conclusions for this work while also proposing some possible future research directions.

\section{Background}
\label{sec:background}
Super resolution is a fundamental subtask of image generation that focuses on increasing the resolution of degraded images. The goal is to enhance the visual quality and level of detail in images that are originally captured or stored at lower resolutions. 
\begin{figure*}[!t]
    \centering
    \includegraphics[width=0.9\linewidth]{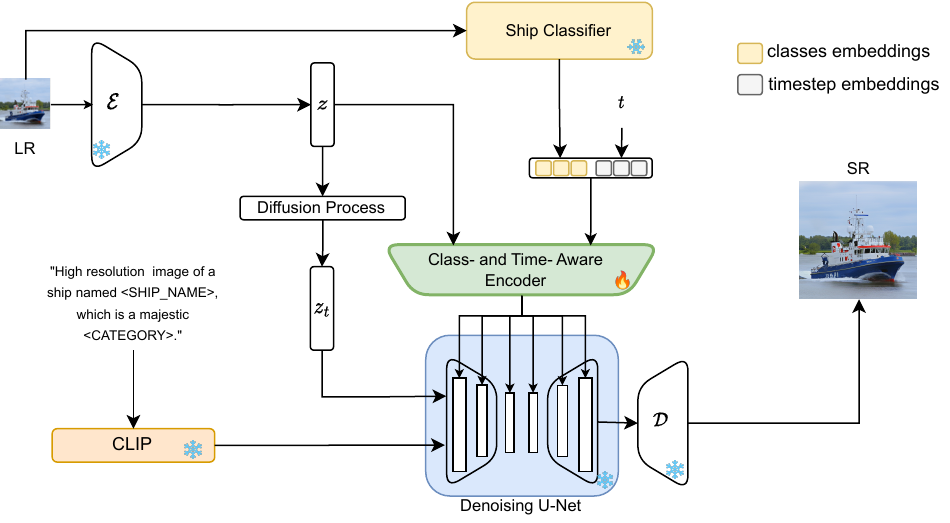}
    \caption{Overview of Stable Ship-Image Super-Resolution framework. We pre-trained the classifier and trained only the class- and time- aware encoder. The other parts of the framework are frozen, which speeds up the overall training process.}
    \label{fig:architecture}
\end{figure*}
In our case, given a $64\times 64$ image, the objective is to generate a higher-resolution output image, such as a $512\times 512$ image, performing an 8x up-scaling, thus laying in the case of Single-Image Super Resolution (SISR).
Our objective is to reconstruct a high-resolution image $\mathbf{x}$ based on a given low-resolution image $\tilde{\mathbf{x}}$. We assume that the relationship between $\tilde{\mathbf{x}}$ and $\mathbf{x}$ can be represented as $\tilde{\mathbf{x}} = (\mathbf{x} \otimes k) + n$, where $k$ denotes the degradation matrix, $\otimes$ represents the Kronecker product that combines the high-resolution image and the degradation matrix, and $n$ represents the noise term. This is the formulation that describes the super-resolution task and serves as the foundation for exploring various algorithms and techniques. 
It is important to note that this problem becomes more challenging due to the existence of multiple potential solutions $x_1, \dots, x_m$ that, when combined with the degradation matrix, can yield the same low-resolution image. More formally, we have a situation where:
\begin{equation}
\tilde{\mathbf{x}} = (\mathbf{x_1} \otimes k) + n = \dots = (\mathbf{x_m} \otimes k) + n
\end{equation}
The presence of such ambiguity poses a significant challenge in real-world scenarios, as it lacks a definitive and singular solution.

Deep learning-based approaches for super resolution have exhibited remarkable advancements over traditional methods, resulting in superior and visually appealing results. Convolutional Neural Networks (CNNs) form the foundation of deep learning techniques employed in super resolution. These networks are capable of learning and extracting pertinent features from low-resolution images, subsequently utilizing these features to generate high-resolution counterparts \cite{Kim2015AccurateIS}. 
Various CNN architectures have been developed for super resolution \cite{dong2015image, ledig2017photorealistic, Zhang2018ImageSU}, which have made significant contributions to the field. Vision Transformer (ViT) introduced in \cite{dosovitskiy2021image} facilitated the successful application of transformers in computer vision. However, the ViT faces challenges when applied to high-resolution images due to fixed-sized patches. To overcome these limitations, the Swin Transformer architecture was introduced in \cite{liu2021swin}.
This model adopts a hierarchical architecture that aggregates information across multiple scales, addressing the fixed patch size constraint of the ViT. It introduces shifting windows, allowing the model to capture intricate details within the image.
SwinIR \cite{swinir}, built upon the previous Swin architecture, is designed for general image restoration excelling in super resolution as well as in other restoration tasks.
SwinIR effectively processes images of large sizes, a characteristic of CNNs, while utilizing a local attention mechanism to handle specific regions of interest within large images. The shifted window design enables SwinIR to effectively model long-range dependencies, capturing global patterns and dependencies in the image, thus leading to enhanced results. A limitation of SwinIR is its potential lack of comprehensive understanding of the global context and semantics of the entire image. While the shifted window technique partially addresses this concern, it still poses constraints on the ability of the model to effectively handle intricate image super-resolution tasks.
However, super-resolution approaches based on deep generative models such as GANs \cite{Menon2020PULSESP} and normalizing flows \cite{10.5555/3495724.3496243} lead to a shift in the super-resolution scenario. Moreover, in recent years, the class of generative models that has garnered considerable attention for a wider range of image generation tasks is the diffusion-based one, which has demonstrated superiority over state-of-the-art methods built using alternative approaches \cite{Dhariwal2021DiffusionMB}. 
The idea of diffusion models was introduced in \cite{sohldickstein2015deep}, but what marked a significant turning point was the work conducted in \cite{Song2020DenoisingDI}, paving the way to a plethora of models involving the creation of new and realistic images that closely resemble samples from a specific domain or adhere to certain constraints \cite{Saharia2022PhotorealisticTD, Song2020DenoisingDI, Podell2023SDXLIL}. 
Those models pervade many different fields even the more delicate like medical imaging \cite{Song2021SolvingIP, Chung2021ScorebasedDM}, generating synthetic data for simulations, and enhancing the quality of imaging techniques. 
It is possible to condition the image generation using external information, providing valuable guidance throughout the generation process, by incorporating additional data or constraints and generating images that align with specific criteria or desired characteristics. For image super resolution, a low-resolution image serves as a guide for a conditioned diffusion process, capable of generating a high-resolution rendition of the original image. 
Among the conditional diffusion models adopted for image super resolution, the SR3 \cite{saharia2021image} model has exhibited remarkable performance. It is based on a conditional diffusion process and utilizes a slightly modified U-Net model as the backbone. Indeed, the authors replace the standard DDPM \cite{Song2020DenoisingDI} Residual Blocks with G-Blocks sourced from \cite{brock2019large}. G-Block is a specific building block within the generator network of BigGAN\cite{brock2019large} that performs upsampling and convolutional operations to enhance the resolution of feature maps. Then, the U-Net takes low-resolution feature maps as input and produces higher-resolution feature maps, enabling the generation of high-resolution images.
However, this approach is both time and computationally expensive, because the diffusion process is done in the pixel space and not in the latent. Nonetheless, our proposed method does not need complete training, instead, it does fine-tuning of a pre-trained text-to-image diffusion model \cite{Rombach2021HighResolutionIS}. Another strategy \cite{9578247} is to exploit the prior knowledge of latent representations given by pre-trained GANs to enhance the quality of super-resolution models. The problem with this approach is the lack of generalizability caused by the limited dataset categories the models were trained on. For this reason, inspired by \cite{wang2023exploiting} we exploit the prior present in large pre-trained diffusion models, that instead are trained with massive image datasets.
\section{StableShip-SR: The Proposed Stable Ship-Image Super-Resolution Method}
\label{sec:method}
Motivated by the research presented in \cite{wang2023exploiting}, our proposed model capitalizes on the foundational knowledge derived from a pre-trained Stable Diffusion model \cite{Rombach2021HighResolutionIS}, based on a U-Net as a denoiser to perform the latent diffusion process. 

A conditional generation process can be directly derived from an unconditional generation one by using direct conditioning in the diffusion process, as done in \cite{saharia2021image}: the latent representation of the degraded input image obtained from an encoder is then used as guidance for the generation process 

Moreover, we exploit the text conditioning given by the design of Stable Diffusion, using CLIP\cite{Radford2021LearningTV} to encode the text and provide embeddings to our model.
We use this representation as an additional conditioning of the denoising process, but only in the training phase. This is necessary to allow the network to be more responsive to the textual information represented by the class and name of the ship, since they are known a priori, being present as metadata in our dataset. This conditioning is not provided in the inference phase, as we want this information to be derived online through the classifier applied to low-resolution images, as analyzed below.
We chose in advance a set of five different parametric prompts to improve the generation process of ship images through the textual information. We randomly pick one of the prompts during the training process and populate it with consistent information about the ship to be processed, i.e. the ship name and the category.

The core of our proposed architecture is the class- and time- aware encoder. Indeed, it provides the diffusion model with additional information on the class of the ship represented in the image to be reconstructed. This information is extracted by a classifier directly from the low-resolution image to be super resoluted.

As a classifier, we utilize a ResNet-50\cite{resnet50} model and finetune it on our proposed dataset ShipSpotting. This pre-training is done before the training of StableShip-SR, consequently, this part of the architecture is frozen during the whole training. 
The classifier training is performed using as input high-resolution images to improve the accuracy of the prediction since it will affect the generation of the images during the denoising process. 

We employ the class- and time- aware encoder output to modulate the intermediate feature maps within the residual blocks of the U-Net through spatial feature transformations (SFT) \cite{8578168}. This slight modification is essential to obtain higher-quality images than other state-of-the-art super-resolution models, as proven in experimental results. 
Following the proposed method in \cite{wang2023exploiting}, we also integrate the temporal information in the encoder, enhancing the overall qualitative result of the generated images. 
Indeed, at early timesteps of the process, when the generated image is still poor in terms of information and consequently the resulting signal-to-noise ratio is low, having stronger conditioning allows the diffusion process to have more helpful guidance. This gives a better understanding of what information needs to be reconstructed, as also in-depth explored in \cite{9879163}. By contrast, at higher timesteps when the diffusion process has already reconstructed much of the image information, i.e., when you have higher signal-to-noise ratio values, the conditioning given by the low-resolution image may be softer, leaving the diffusion process with the finest guidance. Such time awareness is permitted by the fact that the noise schedule is a hyperparameter set beforehand so that the signal-to-noise ratio values at each timestep of the training phase are known. Besides the aforesaid temporal conditioning allows balancing the use of the low-resolution image in the diffusion process. 

We concatenate the two tensors of class and timestep embeddings as shown in Fig. \ref{fig:architecture} resulting in one vector $b \in \mathbb{R}^{1024}$ and then we encode it with the latents of the image $z\in \mathbb{R}^{h\times w\times c}$ obtained by the encoder $z=\mathcal{E}(x)$, where $x\in \mathbb{R}^{3\times H\times W}$ is the low-resolution image. 
The encoder $\mathcal{E}$ and the decoder $\mathcal{D}$ are frozen and are part of the VAE used in \cite{Rombach2021HighResolutionIS}, as well as the neural backbone used to perform the latent diffusion process, i.e. a time-conditional U-Net.
The class- and time- aware encoder $\delta_\theta$ follows the same architecture as the U-Net encoder thus providing conditioning embeddings to the conditional denoising autoencoder $\epsilon_\theta$ at different scales. 
The domain-specific encoder $\tau(y)$, in our case, is a frozen CLIP, where $y$ is the text conditioning. The complete objective function we aim to learn is represented in \eqref{eq:ldm_obj}.
\begin{equation}
\label{eq:ldm_obj}
\small
    L=\mathbb{E}_{\mathcal{E}(x), y, \epsilon \sim \mathcal{N}(0,1), t,c}\left[\left\|\epsilon-\epsilon_\theta\left(z_t, \delta_\theta(c,t,z), \tau(y)\right)\right\|_2^2\right].
\end{equation}
We optimize $\delta_\theta(c,t,z)$ where $(c,t)$ are the classes and the timesteps embeddings respectively, $z$ is the output of the VAE encoder while $z_t$ is its noisy version.



\section{ShipSpotting Dataset}
\label{sec:dataset}
Given the specific task of super resolution for ship images, we create a dataset from scratch, in this way, we can ensure that our super-resolution model is trained on ship images that are relevant and representative. 
It contains ship images that exhibit diversity in terms of ship types, sizes, orientations, and environmental conditions. This diversity is crucial for training an effective super-resolution model capable of handling various real-world scenarios. 
We devote significant attention to constructing a dataset that prioritized generality, ensuring its applicability beyond our specific use case. Our objective is to develop a dataset that does not impose unnecessary restrictions and can be effectively utilized for multiple purposes. The endeavor extended beyond gathering the necessary images for our super-resolution task since we carefully collected additional metadata, making this dataset suitable for other potential problems of interest, such as ship classification.

To construct such a dataset, a straightforward approach was scraping images from the web. The main source for our dataset is ShipSpotting\footnote{\url{www.shipspotting.com}}, which serves as a repository for user-uploaded images, hosting a vast collection of ship images, amounting to approximately $3$ million. Furthermore, for each image, valuable supplementary information is available, such as the type of the ship, and present and past names. 

Next, we made sure that as many images as possible were collected in our dataset, since in deep learning, the quantity of training data directly influences the quality of results.
A larger volume of data enables models to generalize more effectively. Thus we scrape all the images and as a result, the dataset comprises a total of $1.517.702$ samples. 
Motivated by \cite{MARVEL} we exclude many classes of ships from our final analysis and concentrate on the more common and valuable for a real scenario use case. The total number of different classes is $20$ and the ship categories included are Bulkers, Containerships, Cruise ships, Dredgers, Fire Fighting Vessels, Floating Sheerlegs, General Cargo, Inland, Livestock Carriers, Passenger Vessels, Patrol Forces, Reefers, Ro-ro, Supply ships, Tankers, Training ships, Tugs, Vehicle Carriers, Wood Chip Carriers. 
The total amount of samples after this class selection is $507.918$.
To obtain the high-resolution version, we mainly employ a center crop technique, ensuring that the essential details of the image are preserved. Conversely, for the low-resolution counterpart, the high-resolution images were degraded with different techniques following the preprocessing steps given in \cite{9607421}, effectively reducing its resolution.
Furthermore, to create a reference image for comparison with the output of the model, we apply a standard upscaling super-resolution technique to the low-resolution image. The obtained refined dataset is optimized and prepared for training the super-resolution models.

\section{Experimental Results}
\label{sec:experiment}

\begin{figure*}[!t]
\centering
\includegraphics[width=\linewidth]{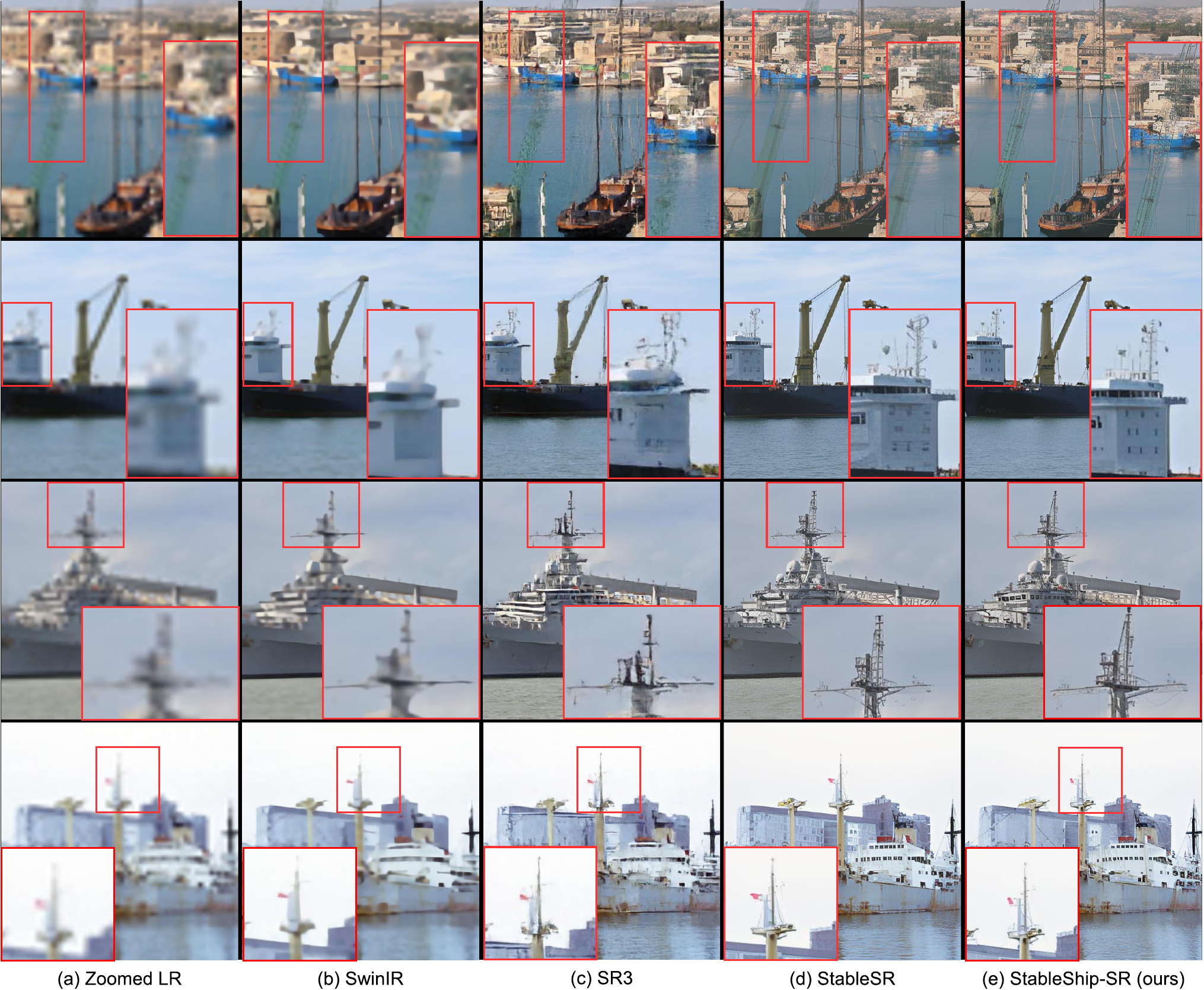}
\caption{Comparison between LR image and SR images reconstructed by other state-of-the-art methods and our proposed model. The red bounding box contains a zoomed-in part of the image in order to have a better understanding of the image resolution quality of our method compared to the others.}
\label{fig:zoomed}
\end{figure*}

We conduct a challenging super-resolution task of 8x scaling, going from $64\times 64$ pixel low-resolution images to $512\times 512$ pixel high-resolution images. We chose this scaling factor to evaluate the ability of the model to handle finer details and accurately recreate them in the output with a very small image as input. 

Models were trained on a machine equipped with an NVIDIA Quadro RTX 8000. After completing the model training, we performed inference on a test set consisting of 1000 images. We analyze the objective metrics PSNR, SSIM, and FID. 
\begin{table}[h!]
\centering
\caption{FID results obtained on the test set.}
\label{tab:objective_results}
\begin{tabular}{|c|c|c|c|}
\hline
Model  & FID $\downarrow$    \\ \hline  
Low Resolution (LR) & 66.05 \\
SwinIR\cite{swinir} & 50.79          \\ 
SR3\cite{saharia2021image}    & 16.59  \\ 
StableSR\cite{wang2023exploiting}    &   12.41  \\
StableShip-SR (Ours)  &  \textbf{11.72} \\ 
\hline
\end{tabular}
\end{table}
A low FID value indicates that the embedding representations of both images are similar, capturing not only general features but also finer details. The images produced by our model are overall more realistic compared to their counterparts, which explains the proximity of their embedding representations, as we can observe from Fig. \ref{fig:zoomed} and results reported in table \ref{tab:objective_results}.
By examining the images, we observe specific characteristics that provide insights into the results. The images generated by Swin-IR and SR3 have a cartoonish appearance, consisting of large areas of color with minimal detail. Each block of color has a well-defined shape, resulting in an artificial aesthetic for the overall image. SwinIR also struggles to accurately reproduce background elements such as water and trees, leading to a slight blurring effect that reduces overall quality.

In contrast, our approach can achieve a better reconstruction of background elements and details, such as portholes and antennas as shown in the third row of Fig. \ref{fig:zoomed}. Moreover, StableShip-SR produces images without the blurred effect present in other approaches. This is likely due to a more realistic approach to color reconstruction, capturing changes in lighting conditions and reflections on water surfaces. It is important to note that the reconstruction of StableSR is not perfect, and careful examination of smaller details reveals artifacts, as we can observe from the first row of \ref{fig:zoomed}.

Another drawback of SR3 is its long inference time, taking $\approx2$ minutes to generate each sample compared to the other models few seconds, this is due to the diffusion process being done on the pixel space and not in the latent one. 

With this analysis we prove StableShip-SR to be a superior model for this specific task, providing higher-quality high-resolution images and more realistic representations compared to the others.

\subsection{Ablation Study on Downstream Tasks}
\label{sec:ablation}
To show the effectiveness and the need for super-resolution images to drastically improve downstream tasks, we conduct experiments in object detection and classification tasks with different images. 

We employ pre-trained neural models in order to have only to fine-tune them on ship datasets. 

Since SeaShip, the ship dataset introduced by \cite{seaships}, also provides bounding boxes, we can test our super-resolution approach in the object detection task, we choose the YOLOv7 \cite{Wang_2023_CVPR} architecture as the object detector, moreover with this model, it is also possible to perform classification tasks. 

Since we are also interested in evaluating the model on our dataset based on ShipSpotting, we fine-tuned the well-known ResNet architecture\cite{resnet50} on ship images as the classifier. 

\begin{figure*}[]
\centering
\includegraphics[width=0.9\linewidth]{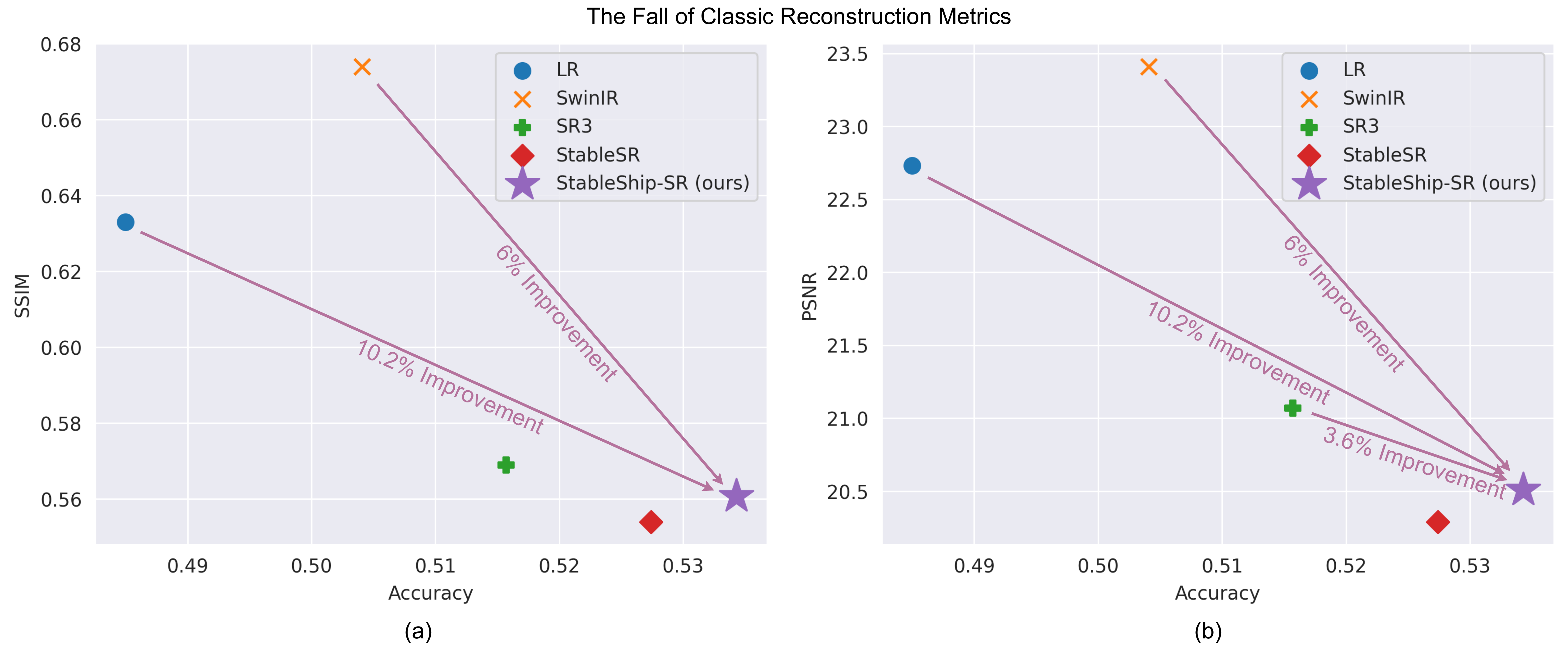}
\caption{Comparison of SSIM (a) and PSNR (b), which measure the quality of the reconstructed images. We confront those results against the classification accuracy reached with the generated images on the ShipsSpotting dataset. This comparison shows how the classical reconstruction metrics are not a measure of how well a model performs in a super-resolution scenario since the accuracy and the FID reached are lower concerning other models that have lower SSIM and PSNR.}
\label{fig:ablation_ShipSpotting}
\end{figure*}

\begin{figure}[]
\centering
\includegraphics[width=0.9\linewidth]{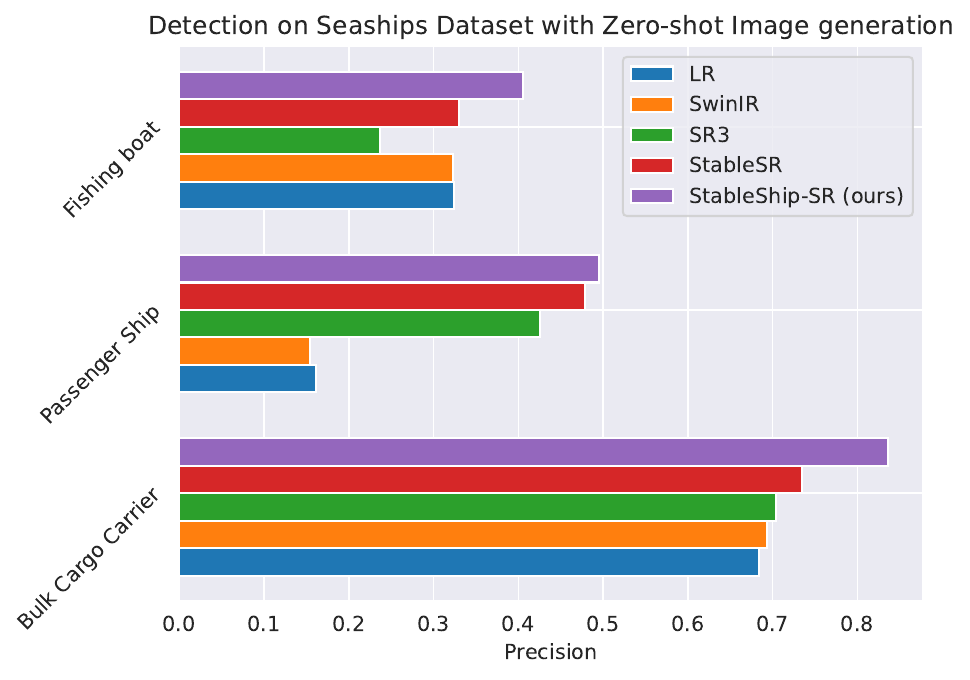}
\caption{Comparison for precision scores on the Seaships dataset. We employ the models pre-trained on our dataset, thus performing zero-shot super resolution, improving the detection of a finetuned YOLOv7\cite{Wang_2023_CVPR} on Seaships.}
\label{fig:bar_chart}
\end{figure}

The results for the detection and classification task performed on the SeaShips dataset are reported in Fig. \ref{fig:bar_chart}. 
SwinIR\cite{swinir} achieves higher PSNR and SSIM values, indicating a higher fidelity between the original and reconstructed images. However, we performed many experiments that point out that those results are useless when it comes to using those images for any other subtasks. Moreover, when considering the FID metric (lower is better), diffusion models outperform SwinIR, and this is outlined also by a subjective evaluation of the results reported in Fig. \ref{fig:zoomed}. These results may seem counterintuitive since we expect that if the original and generated images were similar with low perceptual differences, their embedding representations computed by FID would also be close. However, this is not the case. This behavior aligns with the findings described in \cite{saharia2021image}, where PSNR and SSIM may not accurately represent the quality of a generated image, because they focus on the exact comparison between pixels while generative models may change pixel content without changing their meaning. Indeed the graph in Fig. \ref{fig:ablation_ShipSpotting} shows how a higher PSNR does not bring benefit when it comes to a classification task. Furthermore, using an upsampled low-resolution image as an input for a ship classifier, we will obtain a worse result in terms of accuracy compared to the one obtained with the image with a lower PSNR used in our method. 


\begin{table}[]
\centering
\caption{Overall precision and recall on Seaships dataset. The detection is done by a finetuned YOLOv7\cite{Wang_2023_CVPR} on Seaships.}
\label{tab:precision_recall}
\begin{tabular}{|c|c|c|c|}
\hline
Model  & Precision $\uparrow$           & Recall $\uparrow$    \\  \hline
Low Resolution (LR) & 0.536 &    0.313       \\ 
SwinIR \cite{swinir} & 0.536 & 0.313  \\ 
SR3 \cite{saharia2021image} & 0.569 & 0.316 \\ 
StableSR \cite{wang2023exploiting}    & 0.615  & 0.415 \\
StableShip-SR (Ours)  & \textbf{0.647} & \textbf{0.439} \\ 
\hline
\end{tabular}
\end{table}


\section{Conclusion}
\label{sec:conclusion}
In this paper, we presented StableShip-SR, a state-of-the-art model specifically tailored for ship super resolution. Through comprehensive comparisons across different models, our findings underscore and determine that our method is the most suitable approach for ship super-resolution tasks.
Notably, our model consistently produces images characterized by heightened realism, aligning closely with human perceptual capabilities.

This manuscript delves into the complexities of the super-resolution paradigm from a theoretical standpoint, leveraging a robust architectural foundation. Our experimental evaluations across diverse tasks prove the superiority of StableShip-SR in comparison to its counterparts. Based on our comprehensive testing, we reached some key findings employing both standard and non-standard metrics, evaluating also on downstream tasks to ensure a comprehensive assessment.

A pivotal contribution of our work is the introduction of a meticulously curated ship dataset, containing over $500.000$ samples distributed across $20$ distinct classes.
Being a challenging field of application, we expect that this work will be helpful for the research community as well as for the industry.

Overall, this work mainly contributes to the advancement of research in the field of image super resolution, focusing on the specific application case of ship images, introducing a new model and a new dataset as well as carrying out a performance and trade-off analysis of different approaches.


\bibliography{references}

\begin{thebibliography}{10}
\providecommand{\url}[1]{#1}
\csname url@samestyle\endcsname
\providecommand{\newblock}{\relax}
\providecommand{\bibinfo}[2]{#2}
\providecommand{\BIBentrySTDinterwordspacing}{\spaceskip=0pt\relax}
\providecommand{\BIBentryALTinterwordstretchfactor}{4}
\providecommand{\BIBentryALTinterwordspacing}{\spaceskip=\fontdimen2\font plus
\BIBentryALTinterwordstretchfactor\fontdimen3\font minus
  \fontdimen4\font\relax}
\providecommand{\BIBforeignlanguage}[2]{{%
\expandafter\ifx\csname l@#1\endcsname\relax
\typeout{** WARNING: IEEEtran.bst: No hyphenation pattern has been}%
\typeout{** loaded for the language `#1'. Using the pattern for}%
\typeout{** the default language instead.}%
\else
\language=\csname l@#1\endcsname
\fi
#2}}
\providecommand{\BIBdecl}{\relax}
\BIBdecl

\bibitem{ledig2017photorealistic}
C.~Ledig, L.~Theis, F.~Husz{\'a}r, J.~Caballero, A.~P. Aitken, A.~Tejani,
  J.~Totz, Z.~Wang, and W.~Shi, ``Photo-realistic single image super-resolution
  using a generative adversarial network,'' \emph{IEEE Conf. on Computer Vision
  and Pattern Recognition (CVPR)}, pp. 105--114, 2016.

\bibitem{Menon2020PULSESP}
S.~Menon, A.~Damian, S.~Hu, N.~Ravi, and C.~Rudin, ``Pulse: Self-supervised
  photo upsampling via latent space exploration of generative models,''
  \emph{IEEE/CVF Conf. on Computer Vision and Pattern Recognition (CVPR)}, pp.
  2434--2442, 2020.

\bibitem{Karakus2019ShipWD}
O.~Karakus, I.~G. Rizaev, and A.~M. Achim, ``Ship wake detection in sar images
  via sparse regularization,'' \emph{IEEE Trans. on Geoscience and Remote
  Sensing}, vol.~58, pp. 1665--1677, 2019.

\bibitem{Yang2018PositionDA}
X.~Yang, H.~Sun, X.~Sun, M.~Yan, Z.~Guo, and K.~Fu, ``Position detection and
  direction prediction for arbitrary-oriented ships via multitask rotation
  region convolutional neural network,'' \emph{IEEE Access}, vol.~6, pp.
  50\,839--50\,849, 2018.

\bibitem{Maritime_Vessel_Classification}
C.~Dao-Duc, H.~Xiaohui, and O.~Mor\`{e}re, ``Maritime vessel images
  classification using deep convolutional neural networks,'' in
  \emph{Proceedings of the 6th Int. Symposium on Information and Communication
  Technology}, ser. SoICT '15.\hskip 1em plus 0.5em minus 0.4em\relax New York,
  NY, USA: Association for Computing Machinery, 2015, p. 276–281.

\bibitem{10285968}
L.~Sigillo, A.~Marzilli, D.~Moretti, E.~Grassucci, C.~Greco, and
  D.~Comminiello, ``Sailing the {S}eaformer: A transformer-based model for
  vessel route forecasting,'' in \emph{IEEE 33rd Int. Workshop on Machine
  Learning for Signal Processing (MLSP)}, 2023, pp. 1--6.

\bibitem{seaships}
Z.~Shao, W.~Wu, Z.~Wang, W.~Du, and C.~Li, ``Seaships: A large-scale precisely
  annotated dataset for ship detection,'' \emph{IEEE Trans. on Multimedia},
  vol.~20, no.~10, pp. 2593--2604, 2018.

\bibitem{8455679}
M.~Leclerc, R.~Tharmarasa, M.~C. Florea, A.-C. Boury-Brisset, T.~Kirubarajan,
  and N.~Duclos-Hindié, ``Ship classification using deep learning techniques
  for maritime target tracking,'' in \emph{2018 21st Int. Conf. on Information
  Fusion (FUSION)}, 2018, pp. 737--744.

\bibitem{9024119}
Y.~Shan, X.~Zhou, S.~Liu, Y.~Zhang, and K.~Huang, ``Siam{FPN}: {A} deep
  learning method for accurate and real-time maritime ship tracking,''
  \emph{IEEE Trans. on Circuits and Systems for Video Technology}, vol.~31,
  no.~1, pp. 315--325, 2021.

\bibitem{Song2020DenoisingDI}
J.~Song, C.~Meng, and S.~Ermon, ``Denoising diffusion implicit models,'' in
  \emph{Int. Conf. on Learning Representations}, 2021.

\bibitem{Wang2019DeepLF}
Z.~Wang, J.~Chen, and S.~C.~H. Hoi, ``Deep learning for image super-resolution:
  A survey,'' \emph{IEEE Trans. on Pattern Analysis and Machine Intelligence},
  vol.~43, pp. 3365--3387, 2019.

\bibitem{Wan2020OldPR}
Z.~Wan, B.~Zhang, D.~Chen, P.~Zhang, D.~Chen, J.~Liao, and F.~Wen, ``Old photo
  restoration via deep latent space translation,'' \emph{IEEE Trans. on Pattern
  Analysis and Machine Intelligence}, vol.~45, pp. 2071--2087.

\bibitem{Yeh2016SemanticII}
R.~A. Yeh, C.~Chen, T.-Y. Lim, A.~G. Schwing, M.~A. Hasegawa-Johnson, and M.~N.
  Do, ``Semantic image inpainting with deep generative models,'' \emph{2017
  IEEE Conf. on Computer Vision and Pattern Recognition (CVPR)}, pp.
  6882--6890, 2016.

\bibitem{Ulyanov2017DeepIP}
D.~Ulyanov, A.~Vedaldi, and V.~S. Lempitsky, ``Deep image prior,'' \emph{Int.
  Journal of Computer Vision}, vol. 128, pp. 1867--1888.

\bibitem{Zamir2021RestormerET}
S.~W. Zamir, A.~Arora, S.~H. Khan, M.~Hayat, F.~S. Khan, and M.-H. Yang,
  ``Restormer: Efficient transformer for high-resolution image restoration,''
  \emph{2022 IEEE/CVF Conf. on Computer Vision and Pattern Recognition (CVPR)},
  pp. 5718--5729, 2021.

\bibitem{stawgan}
L.~Sigillo, E.~Grassucci, and D.~Comminiello, ``Staw{GAN}: Structural-aware
  generative adversarial networks for infrared image translation,'' in
  \emph{IEEE Int. Symposium on Circuits and Systems (ISCAS)}, 2023.

\bibitem{Wang2018ESRGANES}
X.~Wang, K.~Yu, S.~Wu, J.~Gu, Y.~Liu, C.~Dong, Y.~Qiao, and C.~Change~Loy,
  ``{ESRGAN}: Enhanced super-resolution generative adversarial networks,'' in
  \emph{Proceedings of the European Conf. on computer vision (ECCV) workshops},
  2018, pp. 0--0.

\bibitem{LI202247}
H.~Li, Y.~Yang, M.~Chang, S.~Chen, H.~Feng, Z.~Xu, Q.~Li, and Y.~Chen,
  ``Srdiff: Single image super-resolution with diffusion probabilistic
  models,'' \emph{Neurocomputing}, vol. 479, pp. 47--59, 2022.

\bibitem{Gao_2023_CVPR}
S.~Gao, X.~Liu, B.~Zeng, S.~Xu, Y.~Li, X.~Luo, J.~Liu, X.~Zhen, and B.~Zhang,
  ``Implicit diffusion models for continuous super-resolution,'' in
  \emph{Proceedings of the IEEE/CVF Conf. on Computer Vision and Pattern
  Recognition (CVPR)}, 2023, pp. 10\,021--10\,030.

\bibitem{10353979}
Y.~Xiao, Q.~Yuan, K.~Jiang, J.~He, X.~Jin, and L.~Zhang, ``{EDiffSR}: An
  efficient diffusion probabilistic model for remote sensing image
  super-resolution,'' \emph{IEEE Trans. on Geoscience and Remote Sensing},
  2024.

\bibitem{pmlr-v162-nichol22a}
A.~Q. Nichol, P.~Dhariwal, A.~Ramesh, P.~Shyam, P.~Mishkin, B.~Mcgrew,
  I.~Sutskever, and M.~Chen, ``{GLIDE}: Towards photorealistic image generation
  and editing with text-guided diffusion models,'' in \emph{Proceedings of the
  39th Int. Conf. on Machine Learning}, K.~Chaudhuri, S.~Jegelka, L.~Song,
  C.~Szepesvari, G.~Niu, and S.~Sabato, Eds., vol. 162, 2022, pp.
  16\,784--16\,804.

\bibitem{Dhariwal2021DiffusionMB}
P.~Dhariwal and A.~Nichol, ``Diffusion models beat {GAN}s on image synthesis,''
  in \emph{Advances in Neural Information Processing Systems}, M.~Ranzato,
  A.~Beygelzimer, Y.~Dauphin, P.~Liang, and J.~W. Vaughan, Eds., vol.~34.\hskip
  1em plus 0.5em minus 0.4em\relax Curran Associates, Inc., 2021, pp.
  8780--8794.

\bibitem{Rombach2021HighResolutionIS}
R.~Rombach, A.~Blattmann, D.~Lorenz, P.~Esser, and B.~Ommer, ``High-resolution
  image synthesis with latent diffusion models,'' \emph{IEEE/CVF Conf. on
  Computer Vision and Pattern Recognition (CVPR)}, pp. 10\,674--10\,685, 2021.

\bibitem{Wang_2023_CVPR}
C.-Y. Wang, A.~Bochkovskiy, and H.-Y.~M. Liao, ``{YOLO}v7: Trainable
  bag-of-freebies sets new state-of-the-art for real-time object detectors,''
  in \emph{Proceedings of the IEEE/CVF Conf. on Computer Vision and Pattern
  Recognition (CVPR)}, June 2023, pp. 7464--7475.

\bibitem{Kim2015AccurateIS}
J.~Kim, J.~K. Lee, and K.~M. Lee, ``Accurate image super-resolution using very
  deep convolutional networks,'' \emph{2016 IEEE Conf. on Computer Vision and
  Pattern Recognition (CVPR)}, pp. 1646--1654, 2015.

\bibitem{dong2015image}
C.~Dong, C.~C. Loy, K.~He, and X.~Tang, ``Image super-resolution using deep
  convolutional networks,'' vol.~38, no.~2, p. 295–307, feb 2016.

\bibitem{Zhang2018ImageSU}
Y.~Zhang, K.~Li, K.~Li, L.~Wang, B.~Zhong, and Y.~R. Fu, ``Image
  super-resolution using very deep residual channel attention networks,'' in
  \emph{European Conf. on Computer Vision}, 2018.

\bibitem{dosovitskiy2021image}
A.~Dosovitskiy, L.~Beyer, A.~Kolesnikov, D.~Weissenborn, X.~Zhai,
  T.~Unterthiner, M.~Dehghani, M.~Minderer, G.~Heigold, S.~Gelly, J.~Uszkoreit,
  and N.~Houlsby, ``An image is worth 16x16 words: Transformers for image
  recognition at scale,'' in \emph{Int. Conf. on Learning Representations},
  2021.

\bibitem{liu2021swin}
Z.~Liu, Y.~Lin, Y.~Cao, H.~Hu, Y.~Wei, Z.~Zhang, S.~Lin, and B.~Guo, ``Swin
  transformer: Hierarchical vision transformer using shifted windows,'' in
  \emph{IEEE/CVF Int. Conf. on Computer Vision (ICCV)}, 2021, pp.
  9992--10\,002.

\bibitem{swinir}
J.~Liang, J.~Cao, G.~Sun, K.~Zhang, L.~Van~Gool, and R.~Timofte, ``Swin{IR}:
  Image restoration using swin transformer,'' in \emph{2021 IEEE/CVF Int. Conf.
  on Computer Vision Workshops (ICCVW)}, 2021.

\bibitem{10.5555/3495724.3496243}
J.~J. Yu, K.~G. Derpanis, and M.~A. Brubaker, ``Wavelet flow: Fast training of
  high resolution normalizing flows,'' in \emph{Proceedings of the 34th Int.
  Conf. on Neural Information Processing Systems}, Red Hook, NY, USA, 2020.

\bibitem{sohldickstein2015deep}
J.~Sohl-Dickstein, E.~Weiss, N.~Maheswaranathan, and S.~Ganguli, ``Deep
  unsupervised learning using nonequilibrium thermodynamics,'' in
  \emph{Proceedings of the 32nd Int. Conf. on Machine Learning}, ser.
  Proceedings of Machine Learning Research, F.~Bach and D.~Blei, Eds.,
  vol.~37.\hskip 1em plus 0.5em minus 0.4em\relax Lille, France: PMLR, 07--09
  Jul 2015, pp. 2256--2265.

\bibitem{Saharia2022PhotorealisticTD}
C.~Saharia, W.~Chan, S.~Saxena, L.~Li, J.~Whang, E.~L. Denton, S.~K.~S.
  Ghasemipour, R.~G. Lopes, B.~Karagol-Ayan, T.~Salimans, J.~Ho, D.~J. Fleet,
  and M.~N. 0002, ``Photorealistic text-to-image diffusion models with deep
  language understanding,'' in \emph{Advances in Neural Information Processing
  Systems 35}, S.~Koyejo, S.~Mohamed, A.~Agarwal, D.~Belgrave, K.~Cho, and
  A.~Oh, Eds., 2022.

\bibitem{Podell2023SDXLIL}
D.~Podell, Z.~English, K.~Lacey, A.~Blattmann, T.~Dockhorn, J.~Muller,
  J.~Penna, and R.~Rombach, ``{SDXL}: Improving latent diffusion models for
  high-resolution image synthesis,'' \emph{ArXiv}, vol. abs/2307.01952, 2023.

\bibitem{Song2021SolvingIP}
Y.~Song, L.~Shen, L.~Xing, and S.~Ermon, ``Solving inverse problems in medical
  imaging with score-based generative models,'' in \emph{Int. Conf. on Learning
  Representations}, 2022.

\bibitem{Chung2021ScorebasedDM}
H.~Chung and J.-C. Ye, ``Score-based diffusion models for accelerated mri,''
  \emph{Medical image analysis}, vol.~80, p. 102479, 2021.

\bibitem{saharia2021image}
C.~Saharia, J.~Ho, W.~Chan, T.~Salimans, D.~J. Fleet, and M.~Norouzi, ``Image
  super-resolution via iterative refinement,'' \emph{IEEE Trans. on Pattern
  Analysis and Machine Intelligence}, vol.~45, no.~4, 2023.

\bibitem{brock2019large}
A.~Brock, J.~Donahue, and K.~Simonyan, ``Large scale {GAN} training for high
  fidelity natural image synthesis,'' in \emph{Int. Conf. on Learning
  Representations}, 2019.

\bibitem{9578247}
T.~Yang, P.~Ren, X.~Xie, and L.~Zhang, ``Gan prior embedded network for blind
  face restoration in the wild,'' in \emph{2021 IEEE/CVF Conf. on Computer
  Vision and Pattern Recognition (CVPR)}, 2021, pp. 672--681.

\bibitem{wang2023exploiting}
J.~Wang, Z.~Yue, S.~Zhou, K.~C. Chan, and C.~C. Loy, ``Exploiting diffusion
  prior for real-world image super-resolution,'' in \emph{arXiv preprint
  arXiv:2305.07015}, 2023.

\bibitem{Radford2021LearningTV}
A.~Radford, J.~W. Kim, C.~Hallacy, A.~Ramesh, G.~Goh, S.~Agarwal, G.~Sastry,
  A.~Askell, P.~Mishkin, J.~Clark, G.~Krueger, and I.~Sutskever, ``Learning
  transferable visual models from natural language supervision,'' in \emph{Int.
  Conf. on Machine Learning}, 2021.

\bibitem{resnet50}
K.~He, X.~Zhang, S.~Ren, and J.~Sun, ``Deep residual learning for image
  recognition,'' in \emph{2016 IEEE Conf. on Computer Vision and Pattern
  Recognition (CVPR)}, 2016, pp. 770--778.

\bibitem{8578168}
X.~Wang, K.~Yu, C.~Dong, and C.~Change~Loy, ``Recovering realistic texture in
  image super-resolution by deep spatial feature transform,'' in \emph{IEEE/CVF
  Conf. on Computer Vision and Pattern Recognition}, 2018.

\bibitem{9879163}
J.~Choi, J.~Lee, C.~Shin, S.~Kim, H.~Kim, and S.~Yoon, ``Perception prioritized
  training of diffusion models,'' in \emph{IEEE/CVF Conf. on Computer Vision
  and Pattern Recognition (CVPR)}, 2022.

\bibitem{MARVEL}
E.~Gundogdu, B.~Solmaz, V.~Y{\"u}cesoy, and A.~Ko{\c{c}}, ``{MARVEL}: A
  large-scale image dataset for maritime vessels,'' in \emph{ACCV}, S.-H. Lai,
  V.~Lepetit, K.~Nishino, and Y.~Sato, Eds.\hskip 1em plus 0.5em minus
  0.4em\relax Springer Int. Publishing, 2017.

\bibitem{9607421}
X.~Wang, L.~Xie, C.~Dong, and Y.~Shan, ``Real-{ESRGAN}: Training real-world
  blind super-resolution with pure synthetic data,'' in \emph{IEEE/CVF Int.
  Conf. on Computer Vision Workshops (ICCVW)}, 2021.

\end{thebibliography}
\bibliographystyle{IEEEtran}

\end{document}